\documentclass{article}
\usepackage{spconf,amsmath,graphicx}
\usepackage{multirow}
\usepackage{subfigure}
\usepackage{multicol}
\usepackage{array}
\usepackage{float}
\usepackage{hyperref}
\usepackage{cite}
\usepackage{amsmath}  

\def\b1{{\mathbf 1}}

\title{BioNet$\colon$ infusing biomarker prior into Global-to-Local network for choroid segmentation in Optical Coherence Tomography images}

\name{Huihong Zhang$^{1,3}$, Jianlong Yang$^{1*}$, Kang Zhou$^{2}$, Zhenjie Chai$^{1,2}$, Jun Cheng$^{1}$, Shenghua Gao${^2}$, Jiang Liu$^{1,4}$ \thanks{Thanks to Ningbo 3315 Innovation team grant; Cixi Institute of Biomedical Engineering, Chinese Academy of Sciences (Y60001RA01, Y80002RA01)}  }

\address{ $^{1}$Cixi Institute of Biomedical Engineering,\\ Ningbo Institute of Materials Technology and Engineering, Chinese Academy of Sciences\\ $^{2}$School of Information Science and Technology, ShanghaiTech University, Shanghai, China\\$^{3}$University of Chinese Academy of Sciences, Beijing, China\\ $^{4}$Department of Computer Science and Engineering,\\ Southern University of Science and Technology, Shenzhen, China }

\begin{document}
%

\maketitle

\begin{abstract}
Choroid is the vascular layer of the eye, which is directly related to the incidence and severity of many ocular diseases. Optical Coherence Tomography (OCT) is capable of imaging both the cross-sectional view of retina and choroid, but the segmentation of the choroid region is challenging because of the fuzzy choroid-sclera interface (CSI). In this paper, we propose a biomarker infused global-to-local network (BioNet) for choroid segmentation, which segments the choroid with higher credibility and robustness. Firstly, our method trains a biomarker prediction network to learn the features of the biomarker. Then a global multi-layers segmentation module is applied to segment the OCT image into 12 layers. Finally, the global multi-layered result and the original OCT image are fed into a local choroid segmentation module to segment the choroid region with the biomarker infused as regularizer. We conducted comparison experiments with the state-of-the-art methods on a dataset (named AROD). The experimental results demonstrate the superiority of our method with $90.77\%$ Dice-index and 6.23 pixels Average-unsigned-surface-detection-error, etc.

\end{abstract}

%



\section{Introduction}

\begin{figure}[t]
\vspace{-0.3cm}
    \centering
    \includegraphics[width=1.8in, height=0.9in]{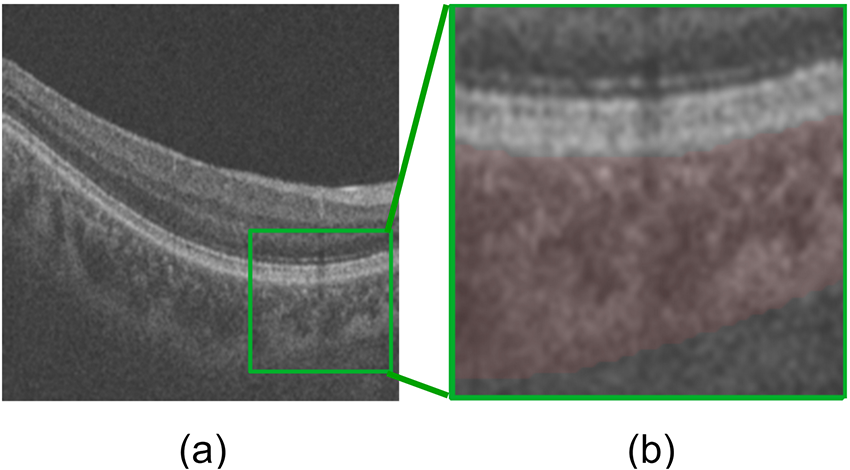}
    \caption{The choroid region in OCT image, where the pink mask represents the choroid region.}
    \label{choroid}
\vspace{-0.5cm}
\end{figure}

Choroid is a vascular structure of the eye, which lies between the retina and the sclera \cite{bio2019}. It provides oxygen and nourishment to the ocular, thus directly related to the incidence and severity of various ocular diseases such as pathologic myopia \cite{myopia2015}, diabetic retinopathy \cite{DR2012}, age-related macular degeneration \cite{AMD2015}, and glaucoma \cite{glaucoma2014}. Optical Coherence Tomography (OCT) is a powerful and non-invasive 3D imaging method to obtain the cross-sectional view of retina and choroid image. However, as shown in Fig. \ref{choroid}, where the pink region indicates the choroid, the lower boundary of the choroid (choroid-sclera interface, referred to as CSI) is quite fuzzy, which makes its segmentation a difficult task. 

Several methods have been proposed for the choroid segmentation in OCT images. Tian \emph{et al.} \cite{Tian:13} presented an automatic algorithm that could segment the choroid region with high accuracy. Alonso \emph{et al.} \cite{Alonso-Caneiro:13} developed an algorithm that detects the upper boundary by an edge filter and a directional weighted map penalty and the lower boundary by OCT image enhancement and a dual brightness probability gradient. Mazzaferri \emph{et al.} \cite{openGraph2017} performed the choroid segmentation via a graph search method.

Recently, with the development of deep learning, the deep neural networks have been widely used in medical image processing. However, there are very few deep learning techniques applied to deal with choroid segmentation tasks. Alonso \emph{et al.} \cite{Alonso2018} tested the performance of two classical deep learning network architecture, Convolutional Neural Network (CNN) and Recurrent Neural Network (RNN), on the retina and choroid boundaries detection task. Masood \emph{et al.} \cite{DeepChoroid2019} adopted a CNN to segment the choroid after a series of pre-processing operations.
\begin{figure*}[t]
\vspace{-0.3cm}
    \centering
    \includegraphics[width=6.2 in, height=2.92 in]{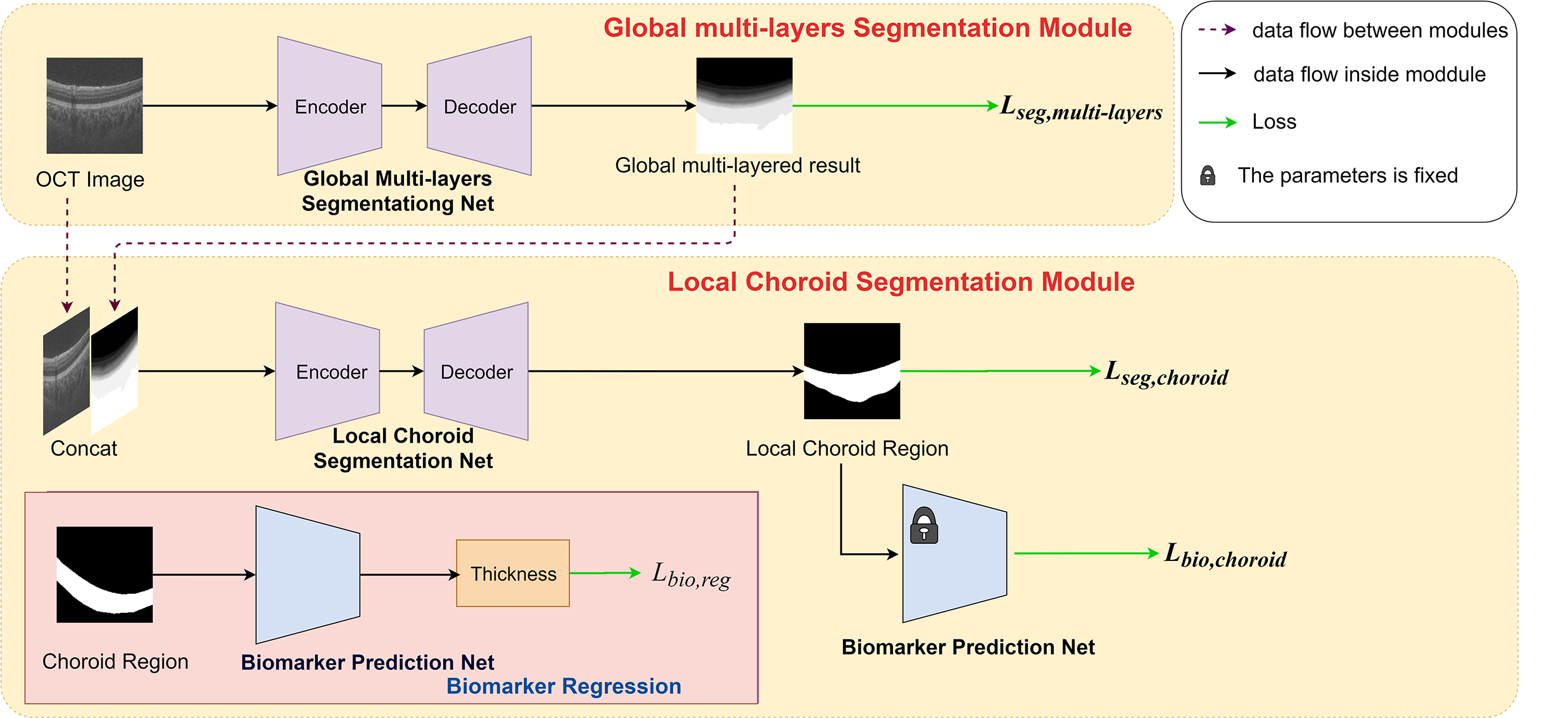}
    \caption{The architecture of our BioNet.}
    \label{framework}
\vspace{-0.5cm}
\end{figure*}

In this paper, we propose a biomarker infused global-to-local network (BioNet) for choroid segmentation in OCT images. Our BioNet contains three parts. Firstly, a biomarker prediction network is trained to predict the biomarker. Then a global segmentation module divides the OCT image into 12 layers. Finally, the multi-layered result is concatenated with the original OCT image to feed into a local choroid segmentation module, which accurately segment the choroid region with the biomarker infused as regularization.

The main contributions of our work include: 1) We propose a biomarker infused global-to-local network (BioNet) for choroid segmentation in OCT images. 2) We combine the global and local information to depress the overfit and improve the robustness of the model. 3) We infuse the biomarker prior into the network to improve the credibility of the segmented choroid. 4) In the experiments, our BioNet obtains the highest performance in choroid segmentation task compared with the state-of-the-art methods.

\section{Motivation of method}

As aforementioned, the choroid is a physiological structure of the human eye, which exists many biomarkers (e.g., thickness, vessel density, etc) in the OCT image \cite{bio2019}. If the biomarker prior knowledge is infused into the network, we are more likely to obtain a result with higher credibility. Among these biomarkers, thickness is a major biomarker for choroid, which denotes the average distance of the upper boundary (Bruce's membrane) and the lower boundary (CSI). As the upper boundary is easy to detect, the thickness regularizer will be helpful to segment the lower boundary. In our BioNet, the biomarker prediction network is trained to predict the thickness of the choroid and later infuse it into the global-to-local network as regularizer.

When a deep learning method is employed to segment the choroid region in OCT image, it is equivalent to a binary classification tasks that divide the pixels into a choroid-class and a non-choroid-class. The model often overfits during training, leading to bad performance on choroid segmentation. That situation may be caused by the small dataset, fuzzy CSI and low local similarity on the non-choroid-class. If we adopt a multi-layer segmentation network to segment the global OCT image into several layers, the similarity on each layer would be higher. Moreover, as a multi-task network, different tasks are constrained with each other, which can reduce overfitting and improve robustness. However, multi-layer segmentation network is trained by a multi-class loss, which aims to optimize the global classification loss. That's to say, the performance of each layer cannot be guaranteed. 


To improve the robustness and achieve better performance, we combine the local choroid segmentation module and the global multi-layers segmentation module to build a global-to-local network for the choroid segmentation. The biomarker is infused to regular the local choroid segmentation, which is helpful to increase the accuracy of the lower boundary.

\section{METHODOLOGY}
\label{method}
In this paper, we propose a biomarker infused global-to-local network for the choroid segmentation task in OCT images. As illustrated in Fig. \ref{framework}, it is a cascade of biomarker prediction network, global multi-layers segmentation module, and local choroid segmentation module. Firstly, the biomarker prediction network is trained to predict the biomarker, and its parameters are fixed after that. Then, we employ the global multi-layers segmentation module to segment the OCT image into 12 layers. Finally, the global multi-layered result and the original OCT image are concatenated and fed into the local choroid segmentation module to segment the choroid region. The biomarker information is infused into the local choroid segmentation module and applied as the regularization.

\begin{figure*}[ttt]
\vspace{-0.3cm}
    \centering
    \includegraphics[width=14cm, height=6cm]{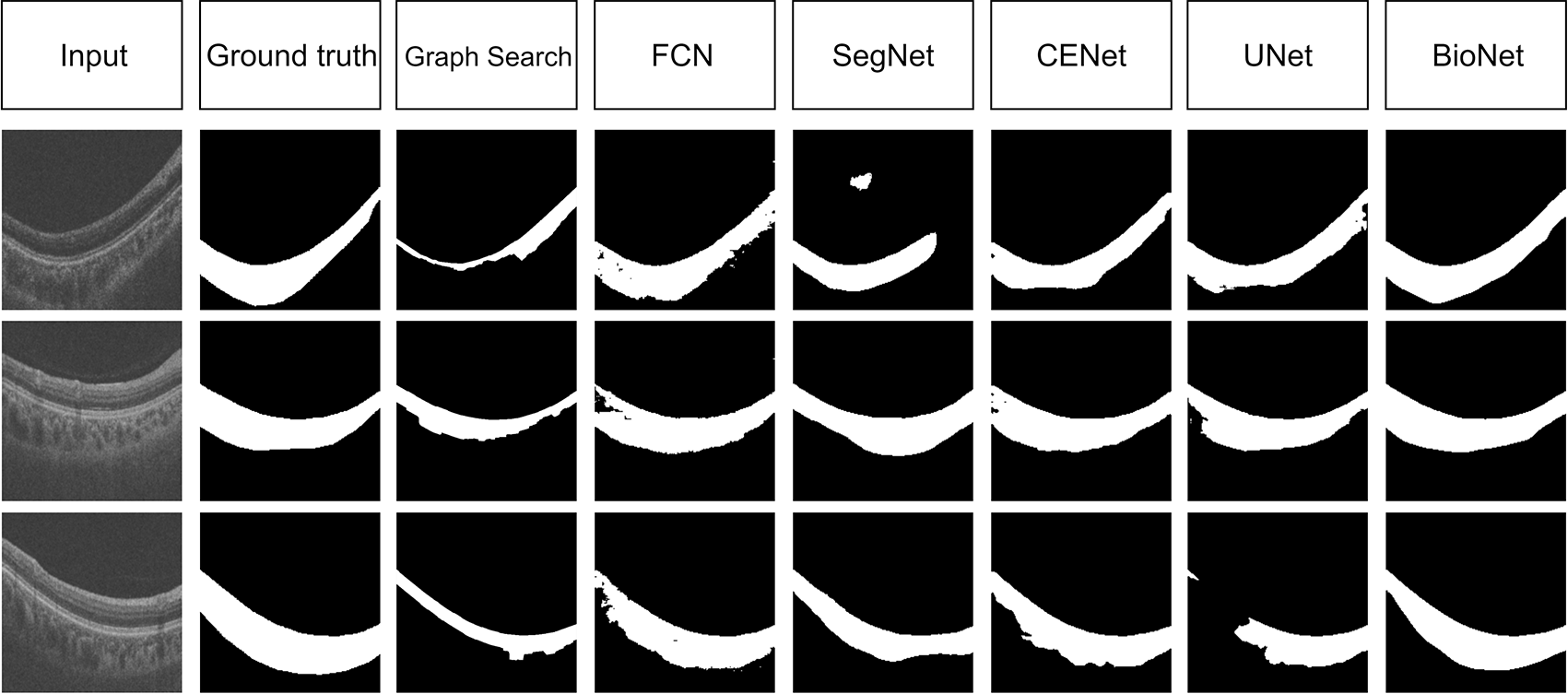}
    \caption{Results of different methods, where the white mask denotes the choroid region.}
    \label{results}
\vspace{-0.3cm}
\end{figure*}

\subsection{Biomarker prediction network:} 
The biomarker prediction network \textbf{B} is trained to predict the biomarker, and the parameters are fixed after that. We utilize the choroidal thickness (an important biomarker of the choroid) in our experiment, which represents the average distance between the upper and lower boundaries. The output of the biomarker prediction network is a predicted biomarker value $B_{pred}$, whose ground-truth is $B_{gt}$. It is trained with mean absolute error loss, denoted as $L_{bio,reg}$.
\begin{equation}
    L_{bio,reg} = \frac{1}{N}\sum_{n=0}^{N-1}||B_{pred}-B_{gt}||
\end{equation}
where $n$ denotes the index of samples, $N$ denotes the total number of samples.

\subsection{Global multi-layers segmentation module:} 
The global multi-layers segmentation module takes OCT image $I_{input}$ as input and estimates the global multi-layers result $G_{pred}$ via a segmentation block $\textbf{U}_{G}$ , $G_{pred}=\textbf{U}_{G}(I_{input})$. For implementation, we adopt U-Net \cite{Ronneberger2015U} as $\textbf{U}_{G}$. $G_{gt}$ denotes the ground truth of the global structure. The $G_{gt}$ is an OCT image with a label of the 12 layers. This module is trained with cross entropy loss. The loss function is calculated as follow:
\begin{equation}
   \begin{aligned}
 L_{seg,multi-layers}=&-\frac{1}{N}\sum_{i=0}^{11}[G_{gt}ln(G_{pred})\\
 & + (1-G_{gt})ln(1-G_{pred})]
 \end{aligned}
\end{equation}
where  $i$ denotes the index of layers.

\subsection{Local choroid segmentation module:}
The local choroid segmentation module takes the concated OCT image and multi-layered result $C_{input}$ as input and predict the choroid region $C_{pred}$ via another segmentation block $\textbf{U}_c$, $C_{pred}=\textbf{U}_c(C_{input})$. It is trained with $L_{seg,choroid}$. When $C_{pred}$ is feed into the biomarker prediction network, it is compared with $B_{pred}$, then another loss termed $L_{bio,choroid}$ is produced.

\begin{equation}
    \begin{aligned}
    L_{seg,choroid}=&-\frac{1}{N}\sum_{i=0}^1[B_{pred}ln(G_{pred})\\
    & + (1-B_{pred})ln(1-G_{pred})]\\
    \end{aligned}
 \end{equation}

\begin{equation}
    L_{bio,choroid} = \frac{1}{N}\sum_{N-1}^{n=0}||\textbf{B}(C_{pred})-B_{pred}||
\end{equation}

Finally, the BioNet is trained by minimizing the total loss.
\begin{equation}
\begin{aligned}
    L_{total} =&\lambda_{seg,multi-layers}L_{seg,multi-layers}\\
    &+ \lambda_{seg,choroid}L_{seg,choroid} + \lambda_{bio,choroid}L_{bio,choroid}
    \end{aligned}
\end{equation}
where $\lambda_{seg,multi-layers}=1$, $\lambda_{seg,choroid}=1$, $\lambda_{bio,choroid}=0.01$, they denote hyper-parameters.

\section{EXPERIMENTAL}
\label{Results}

\subsection{Dataset and Experimental Setup }

In this paper, we use a local dataset named AROD with $256\times20$ B-scans \cite{Gu2019CE}. The B-scan covers a $6\times6$ mm$^2$ region and in a depth of $2$ mm. Each B-scan has $512$ A-lines with $992$ pixels in each A-scan. As the B-scans in the same volume have high similarity, Cheng \emph{et al.} \cite{Cheng2016Speckle} marks the boundary information for $1/4$ of the B-scans, thus $256/4\times20=1280$ B-scans are used. The $1280$ B-scan images were randomly divided into a train set and a test set, each set contains 640 images.

In the experiment, we utilize flipping and rotation to augment the data. The base network is an U-Net \cite{Ronneberger2015U}. The optimizer of our model is Adam \cite{Kingma2014Adam}. The initial value of the learning rate is 0.01, and then the learning rate is respectively reduced to 1/10 of the original when the number of iterations is 40, 80, 160, and 240. The framework is implemented in PyTorch.
\subsection{Evaluation Metrics}
In this paper, we employ the Dice-index (DI), Intersection-over-union (IOU), Average-unsigned -surface-detection-error (AUSDE), Accuracy (Acc), and Sensitivity (Sen) as quantitative metrics. They can evaluate the performances of the methods in different aspects. The DI and IOU mainly show the proportion of the overlap between the segmented choroid region and the ground-truth. The higher the value, the better the performance. AUSDE \cite{Ausde} represents the error of the segmented choroid boundary, and the smaller its value, the better the performance. Acc and Sen are very common parameters that can represent the accuracy and sensitivity of the entire segmented image respectively.

\begin{figure}[t]
\vspace{-0.3cm}
    \centering
    \includegraphics[width=3.4in, height=1.3in]{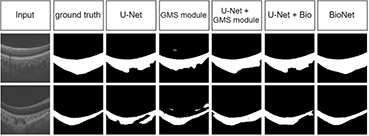}
    \caption{Ablation study. From left to right: Input (OCT image); ground truth of the choroid region; output of U-Net; output of global multi-layers segmentation module; output of U-Net combined with global multi-layers segmentation module; output of U-Net infused by biomarker prediction network; output of our BioNet.}
    \label{ablation}
\vspace{-0.3cm}
\end{figure}

\begin{table}[htb]
 \scalebox{0.86}{
\begin{tabular}{llllll}
\hline
\multicolumn{1}{c}{}                         & \multicolumn{5}{c}{Metrics}                                                                                                        \\ \cline{2-6} 
\multicolumn{1}{c}{\multirow{-2}{*}{Method}} & \multicolumn{1}{c}{IOU} & \multicolumn{1}{c}{AUSDE} & \multicolumn{1}{c}{DI} & \multicolumn{1}{c}{Acc} & \multicolumn{1}{c}{Sen} \\ \hline
Graph search \cite{openGraph2017}                               & 45.75                  & 49.04                    & 62.78                   & 86.31                  & 49.04                  \\
FCN \cite{Shelhamer2014Fully}                                          & 78.30                  & 7.79                     & 87.83                   & 96.12                  & 90.74                  \\
SegNet \cite{ Badrinarayanan2019SegNet}                                       & 75.44                  & 11.69                    & 86.00                   & 95.68                  & 85.81                  \\
CENet \cite{Gu2019CE}                                        & 79.76                  & 7.61                     & 88.74                   & 96.51                  & 89.09                  \\
U-Net \cite{Ronneberger2015U}                                         & 79.14                  & 8.01                     & 88.36                   & 96.30                  & 90.95                  \\ \hline
GMS module                                    & 79.59                  & 10.04                    & 88.62                   & 96.43                  & 89.79                  \\
U-Net+GMS module                               & 81.34                  & 6.54                     & 89.71                   & 96.75                  &\textbf{91.58}                  \\
U-Net+Bio                       & 82.27                  & 6.46                     & 90.27                   & 97.00                  & 89.98                  \\
\textbf{BioNet}                                       & \textbf{83.10}         & \textbf{6.23}            & \textbf{90.77}          & \textbf{97.14}                   & 90.95                  \\ \hline
\multicolumn{6}{l} \textit{units: IOU (\%), AUSDE (pixels), DI (\%), Acc (\%), Sen (\%)}                                                                \\
\multicolumn{6}{l} \textit{GMS module: Global multi-layers segmentation module}  \\
\multicolumn{6}{l} \textit{Bio: Biomarker infusion} 
\end{tabular}}
	\caption{Peformance by various methods.}
    \label{tab}
\end{table}

\subsection{Results}
We compare our BioNet with the state-of-the-art segmentation methods: (1) The graph-Search method in \cite{openGraph2017}. (2) The deep learning methods of FCN \cite{Shelhamer2014Fully}, SegNet \cite{ Badrinarayanan2019SegNet}, U-Net \cite{Ronneberger2015U} and CE-Net\cite{Gu2019CE}. (3) To evaluate the effectiveness of the global multi-layers segmentation module and the biomarker prediction Net, we combine them with the U-Net respectively. The results are reported in Table. \ref{tab}, Fig. \ref{results} and Fig. \ref{ablation}.

\textbf{Comparison results of different methods:}
Table. \ref{tab} summarizes the DI, IOU, AUSDE, Acc, and Sen on the AROD dataset. There are some observations of Table. \ref{tab}: (1) The graph search method does not work well with only $45.75\%$ DI, $49.04$ pixels AUSDE, and $49.04\%$ Sen. (2) The performance of deep learning methods is much higher than the graph search method with the IOU increased by nearly $40\%$, AUSDE decreased by over $40$ pixels, DI increased by nearly $30\%$, and Sen increased by over $40\%$. (3) Our BioNet outperforms the state-of-the-art methods listed in the table with $90.77\%$ DI, $83.10\%$ IOU, $6.23$ pixels AUSDE, and $97.14\%$ Acc. Qualitative results are shown in Fig. \ref{results}. We can see that almost all these methods detect the choroidal upper boundary well. However, except for our BioNet, the methods don't work well on the lower boundary. These results demonstrate that BioNet segment the choroid region with a higher similarity between the ground truth. 

\textbf{Ablation study:}
Table. \ref{tab} and Fig. \ref{ablation} illustrate the effectiveness of the biomarker prediction network and the global multi-layers segmentation module. In the experiment, we take the U-Net as a baseline, the table demonstrates that the infusion of the biomarker prediction network can lead to an improvement on the choroid segmentation task, as the DI increases from $88.36\%$ to $90.27\%$ and the AUSDE decreases from $8.01$ pixels to $6.46$ pixels.
On the other hand, the performance of the U-Net added by the global multi-layers segmentation module makes the IOU increases from $79.14\%$ to $81.34\%$ and AUSDE decreases from $8.01$ pixels to $6.54$ pixels.
Meanwhile, the AUSDE is $3.50$ pixels lower and Sen $1.79\%$ higher than only GMS module employed, which demonstrates that the global-to-local network works better than a single global multi-layers segmentation module or a single local choroid segmentation module.

\section{CONCLUSION}
\label{Coucusion}
In this paper, we propose a biomarker infused global-to-local network, named BioNet, for choroid segmentation with higher robustness and credibility. The choroidal thickness (a major biomarker of the choroid) is used as regularizer in the BioNet and is proved to be effective in improving the choroid segmentation performance. Meanwhile, we make use of both the global multi-layers segmentation module and the local choroid segmentation module to build a global-to-local network, the combination is better than their individual performance. In the experiments, the BioNet outperforms the other state-of-the-art methods. It is expected to provide better assistance to ophthalmologists in OCT image analysis.


%
\bibliographystyle{IEEEbib}
\bibliography{strings}


\end{document}